\documentclass[conference]{IEEEtran}
\IEEEoverridecommandlockouts
\pdfoutput=1
\usepackage{cite}
\usepackage{amsmath,amssymb,amsfonts}
\usepackage{algorithmic}
\usepackage{graphicx}
\usepackage{textcomp}
\usepackage{xcolor}

\usepackage{bm}
\usepackage{comment}
\usepackage{multirow}

\usepackage{pgfplots}

\pgfplotsset{compat=1.18}
\usepackage{subcaption}

\usepackage{fancyhdr}

\usepackage[ruled,vlined,linesnumbered]{algorithm2e}

\def\BibTeX{{\rm B\kern-.05em{\sc i\kern-.025em b}\kern-.08em
    T\kern-.1667em\lower.7ex\hbox{E}\kern-.125emX}}
\begin{document}

\title{Uncertainty-Aware SAR ATR: Defending Against Adversarial Attacks via Bayesian Neural Networks\\
\thanks{
Accepted by the 2024 IEEE Radar Conference.

© 2024 IEEE. Personal use of this material is permitted. Permission from IEEE must be
obtained for all other uses, in any current or future media, including
reprinting/republishing this material for advertising or promotional purposes, creating new
collective works, for resale or redistribution to servers or lists, or reuse of any copyrighted
component of this work in other works.}}

\author{
\IEEEauthorblockN{Tian Ye\IEEEauthorrefmark{1}, Rajgopal Kannan\IEEEauthorrefmark{2}, Viktor Prasanna\IEEEauthorrefmark{1}, Carl Busart\IEEEauthorrefmark{2}}
\IEEEauthorblockA{
    \IEEEauthorrefmark{1}University of Southern California \IEEEauthorrefmark{2}DEVCOM Army Research Lab\\
    \IEEEauthorrefmark{1}\{tye69227, prasanna\}@usc.edu \IEEEauthorrefmark{2}\{rajgopal.kannan.civ, carl.e.busart.civ\}@army.mil}
}

\maketitle

\begin{abstract}
Adversarial attacks have demonstrated the vulnerability of Machine Learning (ML) image classifiers in Synthetic Aperture Radar (SAR) Automatic Target Recognition (ATR) systems. An adversarial attack can deceive the classifier into making incorrect predictions by perturbing the input SAR images, for example, with a few scatterers attached to the on-ground objects. Therefore, it is critical to develop robust SAR ATR systems that can detect potential adversarial attacks by leveraging the inherent uncertainty in ML classifiers, thereby effectively alerting human decision-makers. 
In this paper, we propose a novel uncertainty-aware SAR ATR for detecting adversarial attacks. 
Specifically, we leverage the capability of Bayesian Neural Networks (BNNs) in performing image classification with  \textit{quantified epistemic uncertainty} to measure the confidence for each input SAR image. By evaluating the uncertainty, our method alerts when the input SAR image is likely to be adversarially generated. Simultaneously, we also generate visual explanations that reveal the specific regions in the SAR image where the adversarial scatterers are likely to to be present, thus aiding human decision-making with hints of evidence of adversarial attacks.
Experiments on the MSTAR dataset demonstrate that our approach can identify over 80\% adversarial SAR images with fewer than 20\% false alarms, and our visual explanations can identify up to over 90\% of scatterers in an adversarial SAR image. 

\end{abstract}


\section{Introduction}
\label{sec:intro}

Synthetic Aperture Radar (SAR) Automatic Target Recognition (ATR) is a crucial technique in remote sensing with a wide range of real-world applications. 
SAR images are generated by radar sensors and are passed to the ATR systems for object recognition. SAR image classifiers based on machine learning techniques, e.g., Convolutional Neural Networks (CNNs)~\cite{cnn1,cnn2,cnn3,cnn4} and Graph Neural Networks (GNNs)\cite{gnn}, have demonstrated strong abilities in fast and accurate object recognition in SAR images. 

However, it has also been revealed that these classifiers are vulnerable to adversarial attacks. Carefully crafted noise~\cite{pgd} added to the SAR images may deceive classifiers into making incorrect classification. Particularly, scatterer-based adversarial attacks~\cite{scatter-attack,otsa} can deceive the classifiers by simply attaching a few additional scatterers to the on-ground objects. These perturbations on the SAR images are subtle and imperceptible to human eyes, making it difficult for humans, who rely on SAR ATR to make decisions in critical applications (such as military scenarios),
to notice the adversarial attacks, potentially leading to serious consequences. Therefore, it is necessary to develop techniques that can detect adversarial attacks on SAR images and alert human decision-makers.

In this work, we leverage the uncertainty in ML classifiers and propose a novel uncertainty-aware SAR ATR using Bayesian Neural Networks (BNNs)~\cite{bnn2,bnn3}. The most distinctive feature of BNNs is that their outputs are probabilistic rather than deterministic. 
This allows BNNs to output the classification results along with epistemic uncertainty,
which represents how much knowledge the BNNs have learned about the input at the training time.

By inspecting the epistemic uncertainty, our approach can detect possible adversarial attacks in the input SAR images. Specifically, when a BNN generates high epistemic uncertainty, the input SAR image is more likely an adversarial rather than a benign input, because the classifiers are expected to have learned sufficient knowledge about benign inputs but have limited information about adversarial inputs. With a tolerable False Prediction Rate (FPR), our method can detect adversarial inputs with a high True Prediction Rate (TPR). 

Furthermore, once an input SAR image is identified as adversarial, our method will also generate a visual explanation which shows the possible area where the adversarial perturbations are likely to exist. Specifically, we adapt the Guided Backpropagation (GBP)~\cite{gbp} and propose GBP-BNN, a version compatible with BNNs. The GBP-BNN can provide human experts with the evidence of the attack such that further decisions can be made. 

The concrete contributions of this work are as follows:
\begin{itemize}
    \item We develop a SAR ATR using BNNs to detect adversarial attacks in the input SAR images. As far as we know, we are the first to use BNNs to detect adversarial attacks for SAR ATR. 
    \item We  propose GBP-BNN to generate saliency maps of adversarial SAR images that provide human decision makers with hints on the existence of adversarial perturbations. This method is particularly against the scatterer-based adversarial attacks, which is so far the only type of attacks that can be practically applied by attackers.
    \item Experiments on the MSTAR dataset show that our method effectively distinguishes between benign and adversarial SAR images, achieving TPR of 0.66 to 0.86 at a FPR of 0.1, and 0.81 to 0.92 when the FPR is increased to 0.2. Also, our visual explanation approach successfully identifies 69\% to 94\% of adversarial scatterers, offering a powerful tool for human to identify adversarial attacks in SAR images.
\end{itemize}

\section{Preliminaries and Problem Definition}
\label{sec:Background}

\newcommand{\bs}[1]{\boldsymbol{#1}}
\newcommand{\calD}{{\mathcal{D}}}

\subsection{Bayesian Neural Networks with Uncertainty}
\label{sec:background-bnn}

Bayesian Neural Networks (BNNs)~\cite{bnn2,bnn3} are a type of neural network models that leverage the principles of Bayesian statistics. Different from traditional neural networks where the weights are trained as fixed values, BNNs regard the weights as probabilistic distributions. This enables BNNs to estimate their uncertainty of the outputs. 

\textbf{Supervised training:}
Let $\bs{w}$ be the weights in a neural network and $\calD$ be the training dataset. Training a BNN is the process of updating its beliefs about the weights, represented by probabilistic distributions. These beliefs are initialized as a prior distribution $P(\bs{w})$
. When the BNN is exposed to $\calD$, it revises its beliefs with the new information from the data and updates the prior distribution $P(\bs{w})$ into a posterior distribution $P(\bs{w}|\calD)$ according to Bayes' theorem.

\textbf{Inference:} The classification process in a trained BNN for an input $\bs{x}$ can be formulated as:
\begin{equation}
\label{eq:classify}
P(y|\bs{x},\calD)=
\mathbb{E}_{\bs{w}\sim P(\bs{w}|\calD)}[P(y|\bs{x},\bs{w})]
\approx\frac{1}{T}\sum_{i=1}^TP(y|\bs{x},\bs{w}_i). 
\end{equation}
In this equation, $P(y|\bs{x},\calD)$ is the probability of each class $y$ given input $\bs{x}$ that represents the classification result from the BNN. It is the expectation of output $P(y|\bs{x},\bs{w})$ over all possible weights $\bs{w}$ drawn from $P(\bs{w}|\calD)$. In practice, this integral can be approximated using a Monte Carlo method shown in Equation~\ref{eq:classify}.
Specifically, the inference is conducted $T$ times. Each inference samples weight values $\bs{w}_i$ from the $P(\bs{w}|\calD)$ and generates an output $P(y|\bs{x},\bs{w}_i)$. The final classification result is then obtained by taking the average of all $T$ inferences.

\textbf{Uncertainty:}
The variability in the distribution of the weights inherently allows for the assessment of uncertainty in the classification results. 
In this work, we focus on the \textbf{epistemic uncertainty} that arises due to the model's insufficient knowledge about the true data distribution. It can be quantified by the Mutual Information (MI) between the classification result and the model's weights:
\begin{equation}
\label{eq:uncertainty}
\begin{split}
I(y,\bs{w}|&\bs{x},\calD)=H(P(y|\bs{x},\calD))-\mathbb{E}_{\bs{w}\sim P(\bs{w}|\calD)}[H(P(y|\bs{x},\bs{w}))]\\
&\approx H\left(\frac{1}{T}\sum_{i=1}^TP(y|\bs{x},\bs{w}_i)\right)-\frac{1}{T}\sum_{i=1}^T H(P(y|\bs{x},\bs{w}_i))
\end{split}
\end{equation}
Here, $I(y,\bs{w}|\bs{x},\calD)$ is the MI, $H(P(y|\bs{x},\calD))$ is the predictive entropy and $\mathbb{E}_{P(\bs{w}|\calD)}[H(P(y|\bs{x},\bs{w}))]$ is the expected entropy of the predictions. 
Intuitively, the MI in this context quantifies how much additional information about the output $y$ is provided by $\bs{w}$, after observing the input $\bs{x}$ and considering the information already contained in $\calD$. Therefore, a higher value of MI implies that the variability in $\bs{w}$ greatly influences the uncertainty in $y$. 
It reflects the extent to which our lack of knowledge about the exact values of $\bs{w}$ contributes to the uncertainty in $y$.


In this work, we develop SAR image classifiers based on BNNs. We rely on the epistemic uncertainty as a criterion to determine if the input is adversarial or benign.

\subsection{Adversarial Attacks on SAR Image Classifiers}

Neural networks have been shown vulnerable to adversarial attacks~\cite{fgsm,i-fgsm,mi-fgsm}. For an image classifier $f$ and an input image $\bs{x}$ with ground truth label $y$, an adversarial attack can perturb $\bs{x}$ with carefully designed noise $\bs{\delta}$ and fool the model to make a wrong classification on the perturbed image, i.e., $f(\bs{x}+\bs{\delta})\ne y$. Adversarial attacks also ensure that the perturbation is imperceptible to human eyes.
For example, a traditional adversarial attack method is Projected Gradient Descent (PGD)~\cite{pgd} that iteratively optimizes $\bs{\delta}$ to maximize $L(f(\bs{x}+\bs{\delta}), y)$ with the constraint of $\lVert\bs{\delta}\rVert_p\le\epsilon$ where $\lVert\cdot\rVert_p$ is $L_p$-norm. Several prior works~\cite{sar-attack2,sar-attack4,sar-attack5} apply traditional adversarial attack methods like PGD to the domain of SAR images. However, attacks in this nature requires the attacker to break into the SAR system to precisely manipulate the pixels in SAR images with noise $\bs{\delta}$, which is hardly feasible in real-world scenarios. 

To circumvent this problem, recent studies~\cite{scatter-attack,otsa} have proposed realistic adversarial attacks based on scatterers.
The idea is to perturb the SAR images by putting scatterers to the on-ground vehicles. The scatterers are objects with different scattering structures, coatings, textures, etc than the original vehicle so that they can reflect radar signals differently. 
These attacks design scatterers such that perturbed SAR images perturbed deceive the classifier. It is more feasible because all that the attacker needs to do is to attach the designed scatterers to the on-ground vehicle. 
In this paper, we focus on detecting and interpreting the On-Target Scatterer Attack (OTSA)~\cite{otsa}. 
Table~\ref{table:accuracy} shows the accuracy of three SAR image classifiers with and without the attacks. ``OTSA-$n$" represents using $n$ scatterers to perturb each SAR image. Two examples of OTSA-2 and OTSA-3 are illustrated in the left three columns of Figure~\ref{fig:explain}. Details about the classifiers are in Section~\ref{sec:setup}. 

The perturbation caused by scatterers are imperceptible because SAR images are not easy to interpret by human eyes. This challenge motivates our research in this paper that aims to detect and assist human to interpret these scatterers.

\renewcommand{\arraystretch}{1.2}
\begin{table}
\centering
\begin{tabular}{|c|c|c|c|c|}
\hline
& No Attack & OTSA-1 & OTSA-2 & OTSA-3 \\ \hline
AConvNet & 97.3\% & 47.7\% & 15.5\% & 8.0\% \\ \hline
AlexNet & 97.1\% & 41.7\% & 22.2\% & 17.1\% \\ \hline
LConvNet & 98.4\% & 37.8\% & 15.6\% & 12.8\% \\ \hline
\end{tabular}
\caption{Accuracy of SAR image classifiers on MSTAR without and with adversarial attacks.}
\vspace{-0.5cm}
\label{table:accuracy}
\end{table}

\subsection{Problem Definition}

As SAR ATRs are crucial in real-world applications but are vulnerable to adversarial attacks, it is imperative to detect potential perturbations in SAR images effectively. Human decision-makers who depend on SAR ATRs need to be alerted about any suspicious SAR images to make reliable decisions. In this work, we aim to develop a SAR image classifier, denoted as $f$, that achieves two objectives when presented with an input SAR image $\bs{x}$:
\begin{enumerate}
    \item The classifier $f$ should be able to discern whether the SAR image is benign or adversarial
    . If the image $\bs{x}$ is benign, $f$ should predict the class $y$ of the object depicted in the image with high accuracy.
    \item In cases where the SAR image $\bs{x}$ is identified as adversarial, $f$ should visually indicate the specific locations within the SAR image where adversarial scatterers are likely to have occurred.
\end{enumerate}

\section{Proposed Method}
\label{sec:proposed}

\begin{figure*}
  \centering  \includegraphics[width=\textwidth]{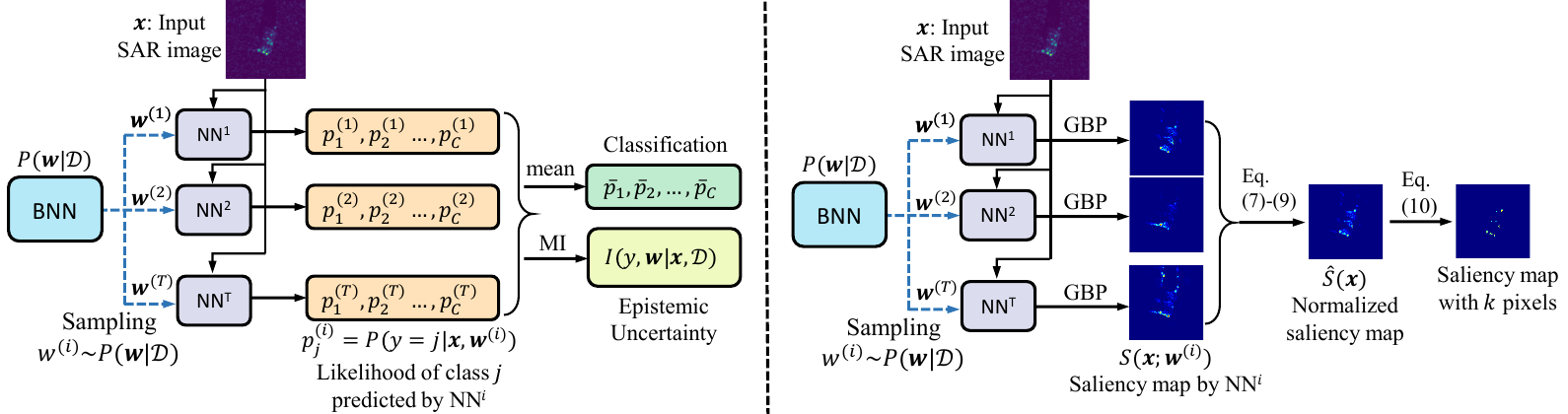}
  \caption{Overview of our methods. Left: Classifying an input SAR image and calculating the epistemic uncertainty. (Algorithm 1 is not illustrated.) Right: Visual explanation of adversarial attacks by GBP-BNN.}
  \label{fig:overview}
  \vspace{-0.4cm}
\end{figure*}

\subsection{Detecting Adversarial Inputs by Epistemic Uncertainty}
\label{sec:method-detect}

The first objective can be well satisfied by choosing BNNs as SAR image classifiers. As described in Section~\ref{sec:background-bnn}, BNNs produce classification results associated with epistemic uncertainty. Inspired by~\cite{bnn-uncertainty}, we claim that the epistemic uncertainty by Equation~\ref{eq:uncertainty} can be a criterion of the presence of adversarial attacks for SAR images with an intuition as follows.
The training process of a neural network fundamentally involves teaching the model to learn the data distribution of SAR images. A benign SAR image is essentially a sample from this distribution. In contrast, adversarial attacks typically fool the neural network by pushing the data point away 
towards regions in the data space where the neural network's knowledge is limited. Such adversarial SAR images are not well-represented in the training dataset, implying that the BNN has not been exposed to comparable data points during its training phase. Consequently, when encountering an adversarial SAR image, the BNN tends to produce higher epistemic uncertainty compared to its response to a benign SAR image.

\setlength{\textfloatsep}{0pt}
\begin{algorithm}
\SetAlgoLined
\DontPrintSemicolon
\SetKwInput{Input}{Input}
\SetKwInput{Output}{Output}
\Input{Trained BNN $f$; Tolerance of FPR $\alpha$; Validation set $V=\{(x_i,c_i)\mid 1\le i\le 2N\}$}
\Output{Threshold $\theta_\text{max}$}

$U\gets \varnothing$, $\textit{TPR}_\text{max}\gets 0$, $\theta_\text{max}\gets 0$

\For{each $(x_i,c_i)$ in $V$} {
$\textit{Use } f \textit{ to generate epistemic uncertainty } u_i$

$U\gets U \cup \{u_i\}$
}

\For{$\theta$ in $U$} {
    $\textit{Compute TPR and FPR on } V \textit{ with }\theta \textit{ as threshold}$

    \If{$\textit{FPR}\le\alpha$ and $\textit{TPR}_\text{max} < \textit{TPR}$} {
        $\textit{TPR}_\text{max}\gets \textit{TPR}$
        
        $\theta_\text{max}\gets \theta$
    }
}

\caption{Find Threshold of Uncertainty}
\label{algo:threshold}
\end{algorithm}
\setlength{\textfloatsep}{1pt}

For an input SAR image $\bs{x}$, we can use BNN to classify it $T$ times, compute the final classification result by Equation~\ref{eq:classify} and calculate the epistemic uncertainty by Equation~\ref{eq:uncertainty}. This procedure is illustrated in the left part of Figure~\ref{fig:overview}. The only thing left is to find a proper threshold $\theta$ such that all SAR images with epistemic uncertainty higher than $\theta$ will be declared as adversarial and otherwise declared as benign. The choice of $\theta$ leads to a tradeoff between the True Positive Rate (TPR) and False Positive Rate (FPR). If we view adversarial SAR images as positive and benign SAR images as negative, then TPR and FPR can be expressed as:
\begin{equation}
\label{eq:tpr}
    \text{TPR}=\frac{\text{\# of adversarial images correctly identified}}{\text{total \# of adversarial images}}
\end{equation}
\begin{equation}
\label{eq:fpr}
    \text{FPR}=\frac{\text{\# of benign images identified as adversarial}}{\text{total \# of benign images}}    
\end{equation}
Raising $\theta$ will make the detection more sensitive to adversarial images, increasing both TPR and FPR, and vice versa.

We allow the human decision-makers, who utilize the SAR ATR, to provide their preference on the tradeoff between TPR and FPR. A preference is in the form of ``\textit{finding the best TPR while keeping FPR no more than $\alpha$}" where $\alpha\in[0,1]$ is the tolerance of FPR. To find the threshold $\theta$, we compose a small set of SAR images with $N$ adversarial SAR images and $N$ benign SAR images as a validation set, denoted as $V=\{(x_i,c_i)\}_{i=1}^{2N}$. In this validation set, $c_i=0$ indicates that $x_i$ is a benign SAR image, and $c_i=1$ indicates that $x_i$ is adversarial. 
Our approach to determine the threshold $\theta$ is presented in Algorithm~\ref{algo:threshold}.
We first compute the epistemic uncertainty for all SAR images in $V$. Then we enumerate all computed uncertainty values as threshold $\theta$ and compute the TPR and FPR over $V$ according to Equations~\ref{eq:tpr} and~\ref{eq:fpr}. The threshold $\theta$ is taken as the one that leads to the highest TPR while satisfying $\text{FPR}\le\alpha$.

\subsection{Visual Explanations of Adversarial Attacks}
\label{sec:method-explain}

While detecting adversarial SAR images provides human decision-makers with an indication of the presence of attacks, it is also important to understand how these images are attacked. 
To this end, we propose to create a visual explanation of the attacks by showing the locations of the scatterers that mislead the classifier to produce wrong classification results. 
This will help decision-makers to 
be fully aware of potential perturbations in the SAR images. Such clarity is particularly vital in scenarios where critical decisions depend on the accuracy and integrity of SAR imagery.

Given that adversarial attacks should be imperceptible to human eyes, OTSA only uses a limited number of scatterers to perturb the SAR images. For the attackers, to significantly impact the classification with just a few scatterers, these scatterers should have high influence to the classification. Therefore, we can create a saliency map for each SAR image that identifies pixels with the most significant contributions to the classification result, and thus the scatterers are highly likely to be spotlighted by the saliency map.
Specifically, we utilize the Guided Backpropagation (GBP)~\cite{gbp} method to create a saliency map that identifies pixels with the most significant contributions to the classification result. 
For a traditional neural network with fixed weights denoted as $\bs{w}$, given an input image $\bs{x}$, GBP produces the saliency map by propagating positive gradients starting from the output layer back to the input layer, denoted as $S(\bs{x};\bs{w})$. 
For BNNs, we adapt GBP to be compatible with probabilistic weights $\bs{w}\sim P(\bs{w}|\calD)$ as follows and call this adapted version \textbf{GBP-BNN}, as illustrated in the right part of Figure~\ref{fig:overview}:
\begin{equation}
\label{eq:S_mean}
    S(\bs{x})=\mathbb{E}_{\bs{w}\sim P(\bs{w}|\calD)}[S(\bs{x};\bs{w})]
\end{equation}
\begin{equation}
\label{eq:std}
    \sigma(\bs{x})=\sqrt{\mathbb{E}_{\bs{w}\sim P(\bs{w}|\calD)}[(S(\bs{x};\bs{w})-S(\bs{x}))^2]}
\end{equation}
where $S(\bs{x})$ and $\sigma(\bs{x})$ denotes the mean and standard deviation of the saliency map. 
As the expectations in Equations~\ref{eq:S_mean} and \ref{eq:std} are intractable to calculate, we can approximate them with Monte Carlo by:
\begin{equation}
    S(\bs{x})\approx\frac{1}{T}\sum_{i=1}^T S(\bs{x};\bs{w}_i)
\end{equation}
\begin{equation}
    \sigma(\bs{x})\approx\sqrt{\frac{1}{T}\left(S(\bs{x};\bs{w}_i)-S(\bs{x})\right)^2}
\end{equation}
Then we normalize the saliency map by the standard deviation:
\begin{equation}
    \hat{S}(\bs{x})=S(\bs{x})/(1+\sigma(\bs{x}))
\end{equation}
We further filter $\hat{S}(\bs{x})$ and only keep the top $k$ pixels:
\begin{equation}
    \hat{S}_k(\bs{x})=\text{Top}_k(\hat{S}(\bs{x}))
\end{equation}
where $\text{Top}_k(\cdot)$ denotes a function retaining the $k$ highest values and setting all other elements to zero. 
This is to limit the number of pixels to highlight so that human can focus exclusively on those pixels with the highest contributions. We claim that the adversarial scatterers are more likely to be located at these pixels. The value of $k$ is set empirically as in Section~\ref{sec:experiments}. If $k$ is too high, more unrelated pixels will be highlighted and make it challenging for human to identify the scatterers. 

It should be noted that GBP-BNN is not applicable to pixel-wise adversarial attacks like PGD, which disperse subtle perturbations across the entire image and make it impossible to spot attacks through individual pixels. However, as stated in Section~\ref{sec:intro}, such attacks are not practical in real-world scenarios. Therefore, excluding visual explanations for these impractical attacks in our study is appropriate.

\section{Experiments}
\label{sec:experiments}

\begin{figure*}
  \centering
  \vspace{-0.25cm}
  \includegraphics[width=0.9\textwidth]{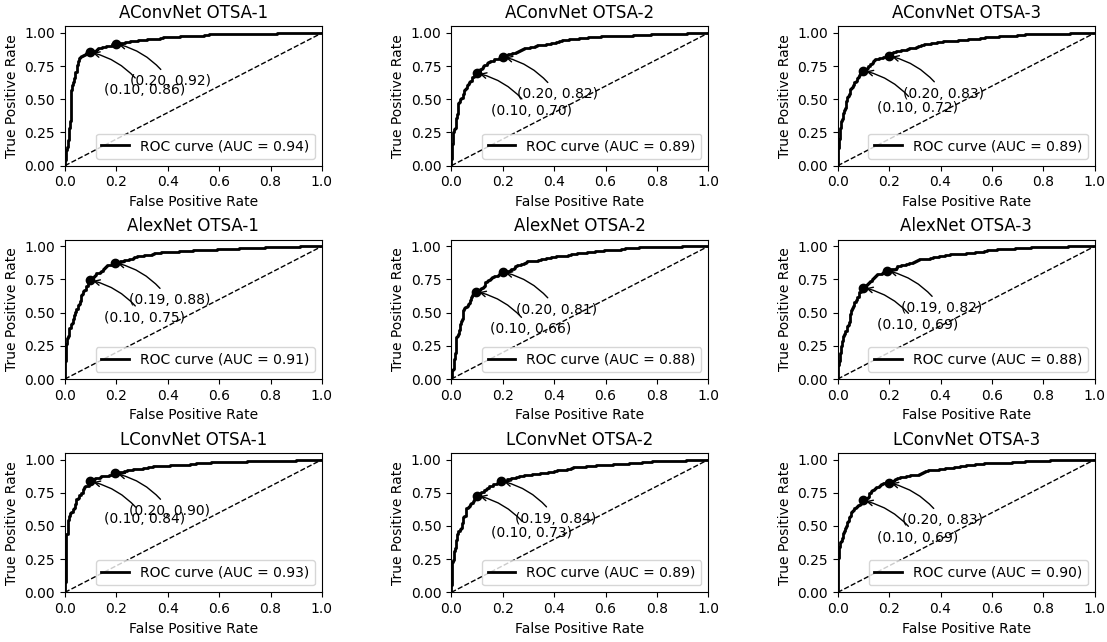}
  \vspace{-0.2cm}
  \caption{ROC curves of detecting adversarial attacks in SAR images by epistemic uncertainty. The three rows from top to bottom are experiments on AConvNet, AlexNet and LConvNet, respectively. The three columns from left to right are for OTSA-1, OTSA-2 and OTSA-3, respectively. In each subfigure, the solid line is the ROC curve for our method while the dashed line is the ROC curve of no discriminative ability. For each ROC curve, two pairs of (FPR, TPR) are shown as specific examples.}
  \label{fig:roc}
  \vspace{-0.35cm}
\end{figure*}

\subsection{Experimental Setup}
\label{sec:setup}

\textbf{Dataset:} Our experiments utilize the MSTAR dataset~\cite{mstar} with SAR images for 10 classes of army vehicles with 2747 SAR images in the training set and 2425 images in the test set. Each image has 128$\times$128 pixels. 
We follow a standard data preprocessing process~\cite{aconvnet}: 
We randomly sample patches of 88$\times$88 pixels from the training set to compose an augmented training set with 27000 images. Each image in the test set are cropped by only keeping the central patch of 88$\times$88 pixels. 

\textbf{Models:} 
Our method is tested on three Bayesian convolutional neural network architectures. For concise representation, we use `C(16,5)' to denote a convolutional layer with 16 output channels and a 5$\times$5 kernel size; `ReLU' represents the Rectified Linear Unit activation function; `MP(2,2)' indicates a max pooling layer with a 2$\times$2 window and stride; and `FC(1024)' refers to a fully connected layer with 1024 neurons. The architectures are detailed as follows:

\begin{itemize}
    \item \textbf{AConvNet: }C(16,5) - ReLU - MP(2,2) - C(32,5) - ReLU - MP(2,2) - C(64,5) - ReLU - MP(2,2) - C(128,6) - ReLU - MP(2,2) - C(10,3)
    \item \textbf{AlexNet: }C(64,11) - ReLU - MP(2,2) - C(192,5) - ReLU - MP(2,2) - C(384,3) - ReLU - C(256,3) - ReLU - C(256,3) - ReLU - MP(2,2) - FC(10)
    \item \textbf{LConvNet: }C(32,5) - ReLU - MP(2,2) - C(64,5) - ReLU - MP(2,2) - C(128,5) - ReLU - MP(2,2) - C(256,6) - ReLU - MP(2,2) - C(256,5) - ReLU - FC(1024) - ReLU - FC(1024) - ReLU - FC(10)
\end{itemize}
The AConvNet architecture was proposed in~\cite{aconvnet} which was designated for the MSTAR dataset. We extend it to a larger network LConvNet and also include AlexNet~\cite{alexnet}
.
The three models are trained with Variational Inference (VI)~\cite{vi}, which is one of the standard methods to train BNNs.

\begin{figure*}
  \centering
  \includegraphics[width=0.7\linewidth]{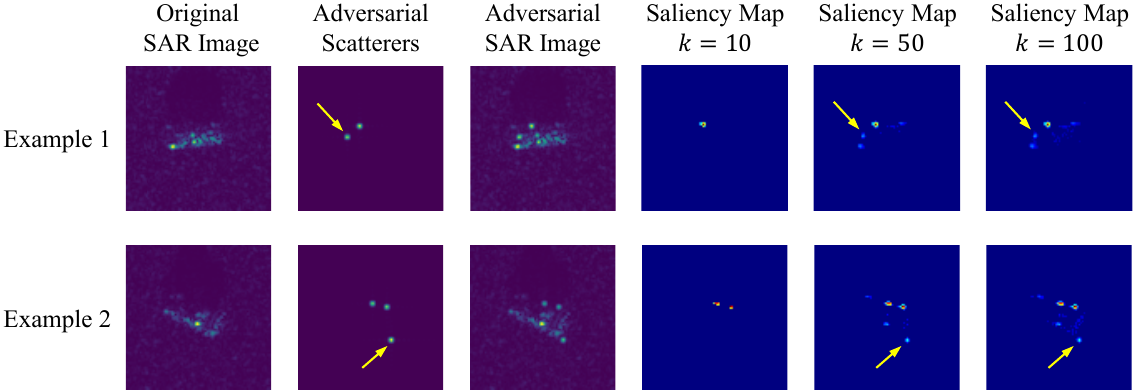}
  \caption{Examples of visual explanation of adversarial attacks. Example 1 is for an OTSA-2 attack on LConvNet. Example 2 is for an OTSA-3 attack on LConvNet.}
  \label{fig:explain}
  \vspace{-0.5cm}
\end{figure*}

\subsection{Metrics}
\label{sec:metrics}
\textbf{Detection of adversarial attacks:} To evaluate the effectiveness of detecting adversarial SAR images with epistemic uncertainty, we plot the Receiver Operating Characteristic (ROC) curve. The ROC curve is a standard metric to evaluate the performance of binary classifications. It shows the tradeoff between TPR and FPR, which is plotted by taking all possible values of the threshold $\theta$. We further use the Area Under the Curve (AUC) ranging from 0 to 1 to evaluate how the detection can distinguish between adversarial and benign SAR images. A higher AUC indicates that the epistemic uncertainty can better separate adversarial and benign SAR images.

\textbf{Visual explanation of adversarial attacks:} To evaluate the performance of the visual explanation of scatterer-based adversarial attacks, we define the metric of Scatterer Identification Ratio (SIR) which can be expressed as $(\text{\# of scatterers identified})/(\text{total \# of scatterers})$. As described in Section~\ref{sec:method-explain}, the saliency map that highlights possible scatterers only keeps $k$ pixels. Clearly, SIR depends on $k$. We use $\text{SIR}_k$ to denote the SIR when the saliency map highlights $k$ pixels.

\subsection{Results for Detection of Adversarial Attacks}

In our experiments, we consider three different OTSA attacks, denoted as OTSA-1, OTSA-2, and OTSA-3, where ``OTSA-$n$" indicates using $n$ scatterers to perturb each image. We utilize 1000 SAR images from the MSTAR test set to generate three distinct datasets. Each dataset consists of 500 benign images and 500 adversarial images perturbed by each OTSA attack, resulting in three datasets corresponding to each attack. Subsequently, 50 benign and 50 adversarial images from each dataset were sampled to determine the epistemic uncertainty threshold $\theta$, as detailed in Algorithm~\ref{algo:threshold}. This process, encompassing all three datasets, was conducted separately on each of the three BNN models, amounting to a total of nine experimental tests.
In Figure~\ref{fig:roc}, we plot the ROC curves and show the corresponding AUCs.
The AUCs of detecting OTSA attacks are consistently near 0.9 across all testing cases, which underscores our method's strong ability of distinguishing benign and adversarial SAR images. The ROC curves exhibit a marked steepness in the lower regions of the FPR.
We also show a few example data points to illustrate the effectiveness of our approach: with the tolerance of FPR set to 0.1, our approach obtains TPRs ranging between 0.66 and 0.86; when the FPR is relaxed to 0.2, our approach achieves TPRs between 0.81 and 0.92. These observations indicate our method's capacity to obtain a high TPR while maintaining a small rate of false alarms.  

\subsection{Results for Visual Explanation of Adversarial Attacks}

For each of the three datasets corresponding to OTSA-1, OTSA-2 and OTSA-3 attacks, we take the 500 adversarial images and test the visual explanation using three SAR image classifiers. Figure~\ref{fig:explain} presents two examples of the visual explanations on OTSA-2 and OTSA-3 attacks applied to LConvNet. 
Each example includes, from left to right, the original SAR image before attacks, the adversarial scatterers, and the adversarial SAR images after attacks. Note that the first two columns are invisible in practical SAR ATR scenarios. The right three columns display saliency maps highlighting 10, 50 and 100 pixels. These maps are tools for human decision-makers to identify adversarial scatterers by comparison with the adversarial SAR images. 
Notably, saliency maps with 50 or 100 pixels can capture all scatterers in both examples, while those with 10 pixels miss one scatterer for each example, as indicated by yellow arrows. Table~\ref{table:explain} details the SIR for 10, 50 and 100 pixels. 
The results show that while saliency maps with 10 pixels identify 23.5\%--73.6\% of scatters, saliency maps with 50 pixels cover 54.9\%--90.7\%. Based on these empirical findings, we suggest setting $k$ to at least 50 to achieve a satisfactory SIR.

\renewcommand{\arraystretch}{1.2}

\begin{table}
\centering
\begin{tabular}{|c|c|c|c|c|}
\hline
& & AConvNet & AlexNet & LConvNet \\ \hline
 & $\text{SIR}_{10}$ & 54.7\% & 63.8\% & 73.6\% \\ \cline{2-5}
OTSA-1 & $\text{SIR}_{50}$ & 81.3\% & 90.0\% & 90.7\% \\ \cline{2-5}
 & $\text{SIR}_{100}$ & 88.7\% & 92.9\% & 94.4\% \\ \hline
 & $\text{SIR}_{10}$ & 37.5\% & 33.5\% & 50.5\% \\ \cline{2-5}
OTSA-2 & $\text{SIR}_{50}$ & 63.9\% & 64.2\% & 77.9\% \\ \cline{2-5}
 & $\text{SIR}_{100}$ & 78.0\% & 77.3\% & 86.3\% \\ \hline
 & $\text{SIR}_{10}$ & 28.0\% & 23.5\% & 36.3\% \\ \cline{2-5}
OTSA-3 & $\text{SIR}_{50}$ & 55.2\% & 54.9\% & 67.0\% \\ \cline{2-5}
 & $\text{SIR}_{100}$ & 68.9\% & 71.3\% & 79.6\% \\ \hline
\end{tabular}
\caption{Scatterer Identification Ratio (SIR) with 10, 50 and 100 pixels highlighted per saliency map.}
\label{table:explain}
\end{table}

\section{Conclusion}
Adversarial attacks can mislead SAR image classifiers by manipulating SAR images with scatterers.
To counter this, we developed SAR image classifiers using BNNs that associate quantified uncertainty with the classification results to detect adversarial attacks in SAR images. We also proposed a visual explanation method to assist human decision-makers in identifying adversarial scatterers. Our methods have shown over 80\% success in detecting adversarial SAR images with a false alarm rate under 20\%. Additionally, our visual explanations can pinpoint up to 90\% of scatterers in SAR images. 
This work opens avenues for further research, such as optimizing model architectures to more effectively distinguish benign and adversarial SAR images, and enhancing visual explanation methods for higher Scatterer Identification Rate (SIR).

\section*{Acknowledgement}
This work is supported by the DEVCOM Army Research Lab (ARL) under grant W911NF2320186.

\bigskip

\textbf{Distribution Statement A:} Approved for public release. Distribution is unlimited.

\bibliographystyle{IEEEtran}
\bibliography{bib}

\end{document}